\documentclass{article}

\usepackage{arxiv}
\usepackage{natbib}
\setcitestyle{authoryear,round,citesep={;},aysep={,},yysep={;}}

\usepackage{amsmath,amsfonts,bm}

\def\figref#1{figure~\ref{#1}}

\def\Figref#1{Figure~\ref{#1}}

\def\secref#1{section~\ref{#1}}

\def\Secref#1{Section~\ref{#1}}

\def\eqref#1{equation~\ref{#1}}

\def\1{\bm{1}}

\DeclareMathAlphabet{\mathsfit}{\encodingdefault}{\sfdefault}{m}{sl}
\SetMathAlphabet{\mathsfit}{bold}{\encodingdefault}{\sfdefault}{bx}{n}

\newcommand{\softmax}{\mathrm{softmax}}
\newcommand{\Enc}{\mathrm{Enc}}

\usepackage{url}
\usepackage{enumitem}
\usepackage{float}
\usepackage{graphicx}
\usepackage{booktabs}
\usepackage{amsthm}
\usepackage{multirow}
\usepackage{caption}
\usepackage{tikz}
\usetikzlibrary{arrows.meta,matrix}
\usepackage[colorlinks=true,citecolor=blue,linkcolor=blue,urlcolor=blue]{hyperref}

\renewcommand{\headeright}{}
\renewcommand{\undertitle}{}
\renewcommand{\shorttitle}{Reasoning on the Simplex: GFPR}

\title{Reasoning on the Simplex: Geometric Fixed-Point Models}

\author{
  {\bf Talgat Daulbaev}$^{3,4,*}$\quad
  {\bf Ilya Glazkov}$^{2,*}$\quad
  {\bf Maxim Rakhuba}$^{2}$\quad
  {\bf Ivan Oseledets}$^{1}$\\[0.6em]
  $^{1}$AXXX \quad
  $^{2}$HSE University \quad
  $^{3}$Applied AI Institute \quad
  $^{4}$LigandPro\\[0.5em]
  {\normalsize $^{*}$Equal contribution}
}

\date{}

\begin{document}

\maketitle

\begin{abstract}
Looped reasoners spend test-time compute by iterating a weight-tied map, but a small residual does not mean the state is a fixed point when that map lives in unconstrained latent space.
We propose Geometric Fixed-Point Reasoning (GFPR), in which the iterated state is the prediction itself: a field of categorical beliefs on a product of simplices, whose $\arg\max$ is the answer at every step.
Because the state is a belief, task structure can be imposed through compact convex relaxations, either as structured readouts or directly in the recurrent state; in the latter case the update remains a continuous self-map, so a fixed point exists for any parameters.
At about 7M parameters, GFPR reaches 95.1\% exact match on Sudoku-Extreme, 92.0\% on Maze-Hard, and 100\% sequence accuracy on $S_5$ length 128, above the published FPRM numbers at the same scale.
The same update also trains a 201M language model on FineWeb-Edu in which each site is a distribution over the vocabulary; with 24 Picard steps it is above GPT-2 small on four zero-shot multiple-choice tasks and above GPT-2 medium on ARC-Easy.
\end{abstract}

\section{Introduction}
\label{sec:intro}

Reasoning benchmarks often mix easy and hard instances, and the hard ones need more internal computation to solve reliably.
A popular response is to add test-time compute without training a larger network, so hard examples can receive more inference steps~\citep{snell2024testtime}.
In large language models, one common mechanism is to generate a longer chain of thought before the final answer~\citep{wei2022cot,kojima2022zeroshot}.
That works well on many open-ended math and logic prompts, but chain-of-thought reasoning is not always the right interface: it ties computation to autoregressive text, grows latency with every token, and offers no grid-shaped state on which to enforce local constraints~\citep{wang2025hrm}.

A different way to spend test-time compute is to apply the same neural block again and again: keep a latent state and update it with a weight-tied map, often for far more steps at inference than during training~\citep{jolicoeurmartineau2025trm,wang2025hrm,movahedi2026fprm,huang2026eqr}.
Read this way, looped reasoning looks like fixed-point iteration: repeat an update until successive states agree, then decode an answer from the settled state~\citep{bai2019deq,movahedi2026fprm,huang2026eqr}.
FPRM~\citep{movahedi2026fprm} makes this explicit and stops when the iterate residual is small, unlike TRM/HRM-style ACT halting~\citep{graves2016act,jolicoeurmartineau2025trm,wang2025hrm}.

Halting on a small residual only certifies an answer if the looped map has a fixed point that the iteration actually reaches.
For an arbitrary latent vector in $\mathbb{R}^d$ the standard guarantee is a contractive map.
\citet{movahedi2026fprm} show that small residual scaling is sufficient for contraction, note that it is not guaranteed in practice, and add a decaying damping step to suppress the resulting oscillations.
On their public Sudoku checkpoint, however, the state at which inference halts is not a fixed point, and even long after the halt the one-step map is typically not locally contractive (\Secref{sec:background}).

We propose Geometric Fixed-Point Reasoning (GFPR), which keeps damped fixed-point iteration but changes the space in which the iterate lives.
The state is a field of categorical beliefs, one probability vector per grid cell or token, and every update returns to this product of simplices through a softmax readout.
After damped Picard iteration the answer is the $\arg\max$ of those beliefs, with known outputs pinned throughout, and the state can be read at any step (\Secref{sec:method}).
The same beliefs also support task-specific convex structure, used either as a structured readout or directly as the recurrent state (\Secref{sec:method-constraints}).
When the recurrent state itself is constrained to a compact convex set, the update remains a continuous self-map, so a fixed point exists; which one the iteration reaches is decided by training (\Secref{sec:method-training}).
At ${\sim}7$M parameters, GFPR reaches 95.1\% exact match on Sudoku-Extreme and 92.0\% on Maze-Hard, compared with 94.2\% and 87.0\% for FPRM at the same scale (\Secref{sec:experiments}).
The same update, trained as a 201M language model on FineWeb-Edu, is compared with GPT-2 on standard zero-shot multiple-choice tasks (\secref{sec:exp-lm}).

To sum up, our contributions are:
\begin{itemize}[leftmargin=1.25em, labelsep=0.35em, topsep=2pt, itemsep=2pt, parsep=0pt]
\item We introduce GFPR, a looped reasoner whose state is a field of categorical beliefs and whose answer is the $\arg\max$ of that state (\secref{sec:method}).
\item We show that this state lets us impose structural constraints for classical reasoning tasks such as Sudoku, mazes, and permutations (\secref{sec:method-constraints}).
\item We evaluate GFPR on Sudoku-Extreme, Maze-Hard, and $S_5$ state tracking, where it is above the published FPRM numbers at similar scale (\secref{sec:experiments}).
\item We train a 201M simplex language model on FineWeb-Edu and compare it with GPT-2 small and medium (\secref{sec:exp-lm}).
\item We study the iteration dynamics of GFPR and compare them with other looped reasoners, showing that GFPR settles to a fixed point while FPRM typically halts at a state that is not one (\secref{sec:background}).
\end{itemize}

\begin{figure}[t]
\centering
\includegraphics[width=\linewidth]{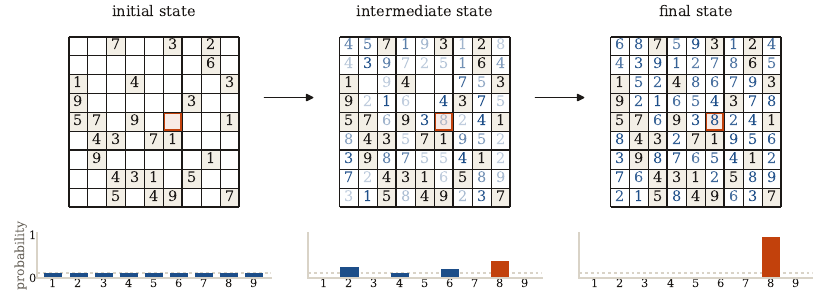}
\caption{In GFPR the state is a field of simplex-valued beliefs: at each site, a distribution over local outputs (e.g.\ Sudoku digits or vocabulary tokens) together with auxiliary register coordinates that store internal state but are not decoded as answers.
At test time we unroll damped Picard steps; the prediction is $\arg\max$ over the output block of the final belief, with known outputs pinned throughout.}
\label{fig:sudoku-settle}
\end{figure}

\section{Background: Looped Reasoning as Fixed-Point Iteration}
\label{sec:background}

A looped reasoner applies one weight-tied map $F_\theta(\cdot\,;x)$ to a state $u$.
With damping $\beta\in(0,1]$,
\begin{equation}
\label{eq:damped-picard}
  u_{k+1} = T_\beta(u_k),
  \qquad
  T_\beta(u) = (1-\beta)\,u + \beta\,F_\theta(u;x),
\end{equation}
whose fixed points satisfy $u^\star=F_\theta(u^\star;x)$.
FPRM~\citep{movahedi2026fprm} halts on a small relative iterate residual with decaying step size; TRM and HRM use ACT or a step cap~\citep{graves2016act,wang2025hrm,jolicoeurmartineau2025trm}; DEQs solve for $u^\star$ implicitly~\citep{bai2019deq}.

\begin{figure}[t]
\centering
\includegraphics[width=\linewidth]{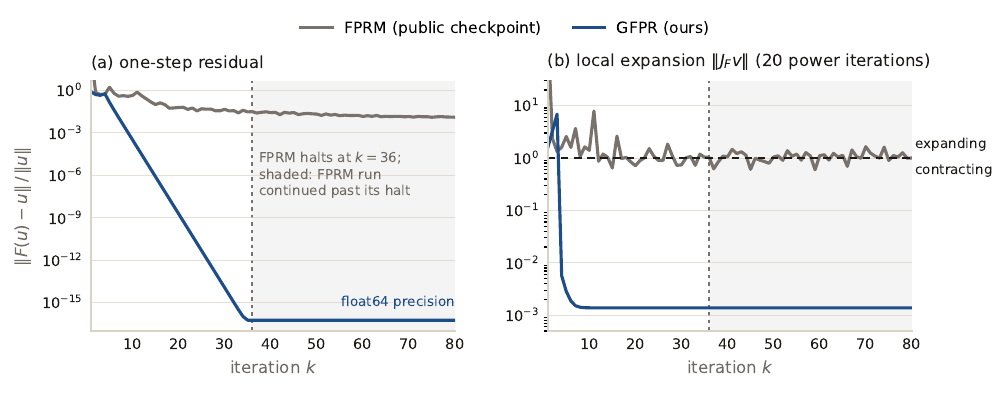}
\caption{One randomly chosen Sudoku-Extreme test puzzle that both FPRM and GFPR solve exactly.
The dotted line marks where FPRM's own rule halts ($k=36$); in the shaded region we keep running FPRM to see whether it settles.
\textbf{(a)}~One-step residual $\|F(u)-u\|/\|u\|$ in float64.
GFPR's residual falls to float64 precision; FPRM's stays near $10^{-2}$ before and after its halt.
\textbf{(b)}~Local amplification probe at each step: $K_{\mathrm{pi}}{=}20$ power iterations on JVPs (float32), plotted as $\|J_F v\|$ after renormalization (not a certified $\rho(J_F)$ when $J_F$ is non-normal; \secref{sec:fprm-protocol}).
The dashed line marks unit amplification.
On the final states of this puzzle only, restarted Arnoldi gives largest Ritz moduli $\approx1.3\times10^{-3}$ (GFPR) and $\approx1.04$ (FPRM).}
\label{fig:fp-residual-compare}
\end{figure}

\paragraph{Existence and reachability.}
Existence of a fixed point and convergence of \eqref{eq:damped-picard} are different questions.
Contraction answers both: if $F_\theta$ is $L$-Lipschitz with $L<1$, Banach's theorem~\citep{banach1922,ortega2000nonlinear,kelley1995iterative} gives a unique fixed point and global convergence, and damping preserves $1-\beta+\beta L<1$.
For a trained network one usually checks local stability at a putative fixed point $u^\star$ through the Jacobian $J_F$: eigenvalues $\lambda$ of $J_F$ become $1-\beta+\beta\lambda$ under damping, and $u^\star$ attracts nearby states when all lie in the unit disk.
If $\rho(J_F)<1$ this holds for every $\beta$; if $\rho(J_F)>1$, $F_\theta$ is not locally contractive in any norm~\citep{horn2012matrix}, yet damped iteration may still converge.
Non-normal $J_F$ can make the residual non-monotone, and a small halt residual need not place the state near $u^\star$ when $J_F$ has an eigenvalue close to one~\citep{trefethen2005spectra,kerg2019nnrnn}.

Brouwer's theorem~\citep{brouwer1911} guarantees a fixed point for every continuous self-map of a nonempty compact convex set, with no contraction and for every $\theta$, but not uniqueness or convergence.
GFPR's belief polytope is of this kind (\secref{sec:method}); training must make the correct equilibrium reachable from the uniform start.

\paragraph{Are FPRM's halted states fixed points?}
FPRM iterates a free latent $y\in\mathbb{R}^d$ and uses pre-norm blocks, residual scaling, and decaying step size $s$ to encourage contraction~\citep{movahedi2026fprm}.
We run its public Sudoku checkpoint with the authors' loop (relative residual threshold $0.1$, $s\leftarrow0.997s$ on stall; supplementary material).

On 1000 held-out Sudoku-Extreme puzzles, 89\% are solved, yet no halted state has relative RMS residual below $10^{-3}$ (median $3.1\times10^{-2}$).
With halting disabled for 8000 steps, the same probe with $K_{\mathrm{pi}}{=}10$ still reports $\|J_F v\|>1$ on 75\% of 128 held-out puzzles (supplementary material); we have not cross-checked that fraction with Arnoldi, so it is a diagnostic of local expansion under the probe rather than a population claim that $\rho(J_F)>1$.

\Figref{fig:fp-residual-compare} is a single-puzzle case study both models solve.
GFPR's residual reaches float64 precision (Arnoldi largest Ritz modulus $\approx1.3\times10^{-3}$ on this puzzle); FPRM's stays near $10^{-2}$ through its halt at $k=36$.
Near the halt the dominant Jacobian mode is the pair $\lambda\approx-0.13\pm1.03i$ ($|\lambda|\approx1.04$).
Any step $s<0.966$ would damp it, but the halt fires at $s\approx0.99$, where $|1-s+s\lambda|\approx1.03$: the update FPRM applies at the state it returns still expands along this mode.

\section{Method}
\label{sec:method}

\subsection{Belief state and update}
\label{sec:method-state}

GFPR represents a problem with $n$ sites (grid cells or tokens) by a state $p=(p_1,\dots,p_n)$, where each $p_i$ is a probability vector over $K$ output symbols and $A$ auxiliary coordinates.
The output block is what the task decodes, for example digits in Sudoku; the auxiliary coordinates act as registers for intermediate computation and are never read out.
Let $\mathcal{K}$ be the set of sites whose output is given by the input (for example Sudoku clues), and let $\mathcal{P}_x$ be the product of simplices in which every site of $\mathcal{K}$ is fixed to the one-hot vector of its given symbol.
With damping $\beta\in(0,1]$, one step is
\begin{equation}
\label{eq:gfpr-step}
  F_\theta(p;\,x) = \Pi_x\Bigl(\softmax\bigl(f_\theta(W_{\mathrm{in}} p + \Enc(x))\bigr)\Bigr), \qquad
  T_\beta(p) = (1-\beta)\, p + \beta\, F_\theta(p;\,x),
\end{equation}
where $f_\theta$ is a Transformer over all sites, the softmax acts per site, and $\Pi_x$ overwrites every site of $\mathcal{K}$ with its one-hot vector.
The prediction at site $i$ is the $\arg\max$ over the output block of the settled $p_i$.
\Figref{fig:sudoku-settle} illustrates this settling process on a Sudoku puzzle.

Since $\mathcal{P}_x$ is nonempty, compact, and convex and $T_\beta$ maps it continuously into itself, a fixed point exists for every $\theta$, $x$, and $\beta\in(0,1]$ (\secref{sec:background}).

\paragraph{Why a simplex.}
Existence alone does not single out the simplex: any continuous map of a compact convex set into itself has a fixed point, so a bounded latent $z\in[-1,1]^d$ updated by $\tanh(f_\theta(z;x))$ has one too.
What the simplex adds is that every intermediate state is itself a prediction.
First, the iteration can be read at any step: the $\arg\max$ of $p_i$ is the current answer at site $i$ and $p_i$ is the model's confidence in it, with no readout head in between.
Second, the residual measures how much the prediction moves.
For any two distributions $|a_j-b_j|\le\mathrm{TV}(a,b)$, so a site whose top two output probabilities differ by more than twice its per-step TV keeps its $\arg\max$ on that step; in a latent chart, a small step in $z$ bounds the change of the decoded answer only through the Lipschitz constant of the head.
Third, the correct solution is a known point of the state space, a vertex of $\mathcal{P}_x$, so the loss pulls the state toward it directly and the distance to it can be tracked along the trajectory.
Fourth, task constraints are statements about beliefs, so they can be imposed on intermediate states rather than checked after decoding (\secref{sec:method-constraints}).

\subsection{Training and inference}
\label{sec:method-training}

Training unrolls the iteration from the uniform state, or with probability $0.25$ from a random Dirichlet state, in two phases.
A first rollout of at most $D$ steps runs without gradients and stops early once every puzzle in the batch has a per-site TV step below a tolerance for two consecutive steps.
From its end state $p^{(0)}$, a tail of $m$ steps $p^{(j+1)}=T_\beta(p^{(j)})$ is differentiated.
With $\tilde p_i$ the renormalized output block of site $i$ at the final state $p^{(m)}$ and $\bar{\mathcal{K}}$ the free sites, the loss is
\begin{equation}
\label{eq:loss}
  \mathcal{L}
  = \frac{1}{|\bar{\mathcal{K}}|}\sum_{i\in\bar{\mathcal{K}}} -\log \tilde p_{i,y_i}
  + \lambda_{\mathrm{aux}}\, a(p^{(m)})
  + \lambda_{\mathrm{res}}\,\frac{1}{m}\sum_{j=0}^{m-1}\frac{1}{n}\sum_{i=1}^{n}\bigl\|\hat F_{\theta,i}(p^{(j)};x)-p^{(j)}_i\bigr\|_2^2,
\end{equation}
where $a(p)$ is the mean auxiliary mass on free sites and $\hat F_\theta=\softmax\bigl(f_\theta(W_{\mathrm{in}} p + \Enc(x))\bigr)$ is $F_\theta$ before $\Pi_x$.
The last term is the one-step residual along the tail; because it is taken before pinning, it also asks the network to reproduce the given sites.
We use $\lambda_{\mathrm{res}}=0.02$ on Sudoku and $\lambda_{\mathrm{res}}=0$ on Maze; depth distributions and other per-task settings are listed in the supplementary material.

\paragraph{Which fixed point is learned.}
Existence does not say which fixed point the test-time loop reaches from the uniform start (\secref{sec:background}).
Training aligns $F_\theta$ with that trajectory: the cross-entropy is taken at the end of a rollout from the same start, so gradients act on the states that inference actually visits rather than on equilibria it may never approach.
The Dirichlet starts ask that the labeled vertex also be reached from a spread of initial beliefs, which widens its basin.

On Sudoku the rollout cap $D$ is redrawn every batch from a heavy-tailed distribution (Maze uses a fixed $D$), so the same weights must give the right answer after short and long rollouts.
A map that passes through the solution at one depth and then moves on is penalized at the others, so the answer has to persist.
Short draws also end the rollout before the state settles, so the loss sees intermediate states as well; training only at settled states would leave the map free to form confident wrong attractors in regions the loss never visits.

With $\lambda_{\mathrm{res}}>0$ on Sudoku, the residual term additionally asks the state the model settles on to be stationary, not merely correct at the end of the tail.
On Maze, $\lambda_{\mathrm{res}}=0$ and only the final answer is supervised.
Near a fixed point, where the Jacobian varies slowly along the tail, backpropagation through the $m$ tail steps approaches a truncated Neumann approximation of the implicit fixed-point gradient, which involves $(I-J_F)^{-1}$~\citep{bai2019deq}.

None of this certifies convergence at test time.
Training rollouts are tens of steps long (mean $D\approx32$ on Sudoku), while test budgets reach $35{,}000$ steps, so behavior beyond the training horizon is extrapolation; a few held-out Sudoku puzzles settle to a stable but incorrect grid.

At test time GFPR starts from the uniform state with the given sites pinned, applies $p_{k+1}=T_\beta(p_k)$ for $K$ steps (on $S_5$, until the total-variation step falls below a tolerance; budgets in the supplementary material), and reads $\arg\max$ on the output block of $p_K$.
No labels, verifier, or restarts are used at test time.

\subsection{Language models: wavefront decode}
\label{sec:method-lm}

The same state applies to a token sequence.
Site $t$ holds a belief $p_t$ over the vocabulary, and training fits the output block of $p_t$ to the next token $x_{t+1}$, with the embedding of the true $x_t$ as conditioning.
One Picard step is one causal pass of $f_\theta$.
On Sudoku and mazes every free site is updated for the full inference depth and the whole answer is read at the end (\secref{sec:method-training}).
Under causal attention the update at site $t$ depends only on sites $\le t$, so $p_t$ is a prediction of $x_{t+1}$ only after $x_1,\ldots,x_t$ are fixed.
Open-ended generation therefore alternates Picard steps with commits, and we use \emph{wavefront} decode for it.
Wavefront decode is purely an inference procedure: the loss \eqref{eq:loss} is teacher-forced, so $F_\theta$ is never trained on an expected embedding.
The multiple-choice scores of \secref{sec:exp-lm} need no generation and are computed from teacher-forced beliefs.

Wavefront decode writes the sequence left to right.
It tracks a frontier $f$: the position of the next token that is not yet fixed.
When the belief $p_{f-1}$ at the site before the frontier looks ready, we commit $x_f=\arg\max p_{f-1}$, pin that site to a one-hot vector, and move $f$ forward by one.
While we wait for readiness, we do not freeze the entire suffix: up to $W$ sites ahead of $f$ (the soft-ahead width) keep taking Picard steps so later beliefs can start to form before the prefix is complete.

Embeddings follow the same split.
Sites that are already committed feed the network their true token vectors.
Inside the window, site $i$ conditions on a soft embedding built from the belief at $i-1$,
\begin{equation}
\label{eq:soft-emb}
  e_i = \sum_w q_{i-1,w}\,\mathrm{Emb}(w),
  \qquad
  q_{i-1} = \softmax(\log p_{i-1}/\tau),
\end{equation}
where $q_{i-1}$ is a temperature-sharpened version of $p_{i-1}$.
Sites beyond the window are not updated until the frontier moves and they enter the window.
On the very first segment, the given prompt is run with teacher-forced embeddings only, so the readiness test at $f$ is not applied to beliefs that are still essentially uniform from the cold start.

Readiness of $p_{f-1}$ is a gap between its top two probabilities, a total-variation residual that has stayed small for several steps, or the per-token step cap.
A minimum number of steps is required after every commit.
The first steps after a commit are small even when the state is far from equilibrium, and a raw residual test treats that pause as convergence.
Pinning $p_{f-1}$ to the committed one-hot makes the soft embedding at the next site match the ordinary token embedding of $x_f$.

Because the update at site $t$ ignores sites $>t$, prefix order is the order the dependencies allow.
The soft window lets up to $W$ later beliefs move before their prefix is frozen.
A margin test can commit a sharp belief that is still moving; a residual test can wait out a belief that has already converged to a flat distribution.
Default $W$, tolerances, and the implementation are in \secref{sec:supp-lm-decode}.

\subsection{Task-specific convex constraints}
\label{sec:method-constraints}

The product-of-simplices state does not couple sites. We encode global task structure by a compact convex relaxation $\mathcal{C}(x)$: a terminal structured readout for Sudoku and Maze-Hard, and the recurrent state for $S_5$. Let $q$ denote the structured answer variable, distinct from the state $p$ of \secref{sec:method-state}. For nonempty compact convex $\mathcal{C}(x)\subset\mathbb{R}^d$ and a continuous convex regularizer $\Omega$, define
\begin{equation}
\label{eq:general-convex-step}
\begin{aligned}
  \Phi_x(z)
  &= \max_{q\in\mathcal{C}(x)}
  \left\{\langle z,q\rangle-\Omega(q)\right\},\\
  \Psi_{\mathcal{C}(x)}(z)
  &= \operatorname*{arg\,max}_{q\in\mathcal{C}(x)}
  \left\{\langle z,q\rangle-\Omega(q)\right\}.
\end{aligned}
\end{equation}
Here $\Psi_{\mathcal{C}(x)}$ is a structured analogue of softmax: it maps unconstrained scores to an answer state satisfying the convex task constraints. If the maximizer is unique, compactness and continuity make $\Psi_{\mathcal{C}(x)}$ continuous in $z$.

When the structured variable $q$ is used as the recurrent state, we update it as
\begin{equation}
\label{eq:structured-recurrence}
  T_{\beta,\mathcal{C}}(q;x)
  = (1-\beta)q
  + \beta\,\Psi_{\mathcal{C}(x)}\bigl(f_\theta(q,x)\bigr),
  \qquad \beta\in(0,1].
\end{equation}
Both terms lie in $\mathcal{C}(x)$, so convexity keeps every update in $\mathcal{C}(x)$. Thus $T_{\beta,\mathcal{C}}$ is a continuous self-map, and the fixed-point existence argument of \secref{sec:background} applies.

\paragraph{Instantiations.}
For Sudoku, $q_{rcd}$ is the belief that cell $(r,c)$ takes digit $d$. The relaxation $\mathcal{C}_{\mathrm{Sud}}(x)$ requires a distribution on each cell, unit total mass of each digit in every row, column, and $3\times3$ box, and $q_{rc,d_{rc}}=1$ on clues (\eqref{eq:sudoku-polytope}). Its points may be fractional, so membership does not itself certify a valid grid. We use the power regularizer $\Omega_\alpha(q)=\frac{1}{\alpha(\alpha-1)}\sum_i q_i^\alpha$ with $\alpha>1$ and evaluate $\Psi_{\mathcal{C}_{\mathrm{Sud}}(x)}$ through its dual (\secref{sec:supp-constraints}).

For Maze-Hard, parent pointers induce a unit flow from the goal to the start on directed grid edges. With incidence matrix $B_x$ and right-hand side $b_x$ (source at the goal, sink at the start),
\begin{equation}
\label{eq:maze-flow-polytope}
  \mathcal{C}_{\mathrm{flow}}(x)
  = \left\{f\in[0,1]^{|E_x|}: B_x f=b_x\right\}.
\end{equation}
Flow conservation does not enforce a simple or shortest path; we decode the resulting parent field and evaluate the route separately. 

For $S_5$, the structured state is a doubly stochastic matrix
$X\in\mathcal{B}_5$, parameterized as
$X=\sum_{\pi\in S_5} w_\pi P_\pi$ over the 120 permutation matrices $P_\pi$.
The weights $w_\pi$ are obtained by applying either softmax or $\alpha$-entmax~\citep{peters2019sparse} to the permutation scores.

\paragraph{Fenchel--Young readout.}
For a target $y\in\mathcal{C}(x)$ we use the Fenchel--Young loss~\citep{blondel2020fenchel}
\begin{equation}
\label{eq:fy-structured}
  L_{\Omega,\mathcal{C}}(z,y)
  = \Phi_x(z)+\Omega(y)-\langle z,y\rangle;
\end{equation}
when the maximizer in \eqref{eq:general-convex-step} is unique,
$\nabla_z L_{\Omega,\mathcal{C}}(z,y)=\Psi_{\mathcal{C}(x)}(z)-y$.
In Table~\ref{tab:main}, Sudoku and Maze retain the original recurrent state $p$, including its auxiliary coordinates, and apply $\Psi_{\mathcal{C}(x)}$ only to the final scores. Thus the structured solver is a terminal readout rather than part of the recurrent loop. The constrained $S_5$ model instead uses the doubly stochastic matrix $X$ as its recurrent state and applies the structured update at every step. Solver details and the entmax $\alpha$ ablation are in \secref{sec:supp-constraints}.
\section{Experiments}
\label{sec:experiments}

\subsection{Tasks}
\label{sec:exp-tasks}

We evaluate on three structured benchmarks used by recent looped reasoners~\citep{wang2025hrm,jolicoeurmartineau2025trm,movahedi2026fprm}, all scored by exact match of the full answer, and on zero-shot language-model multiple choice.

\paragraph{Sudoku-Extreme.}
$9\times9$ puzzles with unique solutions, selected to be hard~\citep{wang2025hrm}.
Training uses the 1{,}000-puzzle split with validity-preserving augmentations (digit relabeling, band and stack permutations, transposition).
Checkpoints are chosen on a stratified 1{,}000-puzzle validation slice of the \texttt{test\_hard} file in \texttt{sapientinc/sudoku-extreme-1k}; the headline number is on the separate public file \texttt{sapientinc/sudoku-extreme} \texttt{test.csv} ($422{,}786$ puzzles) used by HRM and FPRM, which is not used for model selection (\secref{sec:eval}).
Headline evaluation matches the released FPRM Sudoku-Extreme setup~\citep{movahedi2026fprm}: one trajectory from the uniform state, damping $\beta=0.7$, and at most $35{,}000$ damped Picard steps (the same step cap as FPRM on this benchmark).
Each cell is a site with $K=9$ digits, and the clues are pinned.

\paragraph{Maze-Hard.}
$30\times30$ mazes whose shortest path is longer than 110 steps, with 1{,}000 training and 1{,}000 test mazes~\citep{wang2025hrm}.
Rather than predicting the route mask, GFPR predicts at every open cell the direction to its parent in a breadth-first tree rooted at the start; the route is recovered by following parent pointers from the goal and compared with the labeled route.
We train with the eight dihedral symmetries of the square, which the parent field respects because it is recomputed from the transformed maze.
We evaluate on all 1{,}000 test mazes, as in prior work on this benchmark.

\paragraph{$S_5$ state tracking.}
Each instance is an initial arrangement of five elements followed by a sequence of permutations from the symmetric group $S_5$, and the model must output the final arrangement~\citep{merrill2024illusion}.
Following \citet{movahedi2026fprm}, models are trained on sequences of up to 32 updates and evaluated on up to 128, which tests length generalization.
The reported quantity is exact accuracy of the final arrangement.
We enumerate the $120$ permutations in lexicographic order and represent each input permutation and each target prefix product by its index.
GFPR encodes every step as a site on the product of simplices over these group elements, with no coupling across sites, and scores a sequence correct only when every prefix is correct.
The constrained variant keeps a doubly stochastic state $X=\sum_{\pi}w_\pi P_\pi$, where $P_\pi$ are the 120 permutation matrices and the vertex weights $w$ are a softmax of the scores $\langle Z,P_\pi\rangle$; the entmax variant replaces this softmax by $\alpha$-entmax.

\paragraph{Language modeling.}
The same product-of-simplices state is trained as a causal language model on 6B tokens of FineWeb-Edu~\citep{penedo2024fineweb} (\secref{sec:method-lm}).
The released run has 201M parameters (width 2048, 16 heads, two weight-tied blocks, vocabulary 16{,}384 plus 16 auxiliary coordinates, context 512).
We score four zero-shot multiple-choice tasks with next-token log-likelihood, not generated text: ARC-Easy~\citep{clark2018arc}, SciQ~\citep{welbl2017sciq}, PIQA~\citep{bisk2020piqa}, and HellaSwag~\citep{zellers2019hellaswag}.
GPT-2 small (124M) and GPT-2 medium (355M)~\citep{radford2019language} are the comparison points.

\subsection{Baselines}
\label{sec:exp-baselines}

We compare with looped reasoners at a similar parameter scale: HRM~\citep{wang2025hrm}, TRM~\citep{jolicoeurmartineau2025trm}, EqR~\citep{huang2026eqr}, and FPRM~\citep{movahedi2026fprm}.
For Sudoku-Extreme and Maze-Hard, every model in Table~\ref{tab:main} is reported on the same public test files: \texttt{sapientinc/sudoku-extreme} \texttt{test.csv} ($N{=}422{,}786$) and \texttt{sapientinc/maze-30x30-hard-1k} \texttt{test.csv} ($N{=}1{,}000$), as in HRM/TRM/FPRM~\citep{wang2025hrm,jolicoeurmartineau2025trm,movahedi2026fprm}.
We do not re-run the baselines; quoted accuracies come from Table~1 of \citet{movahedi2026fprm}, with TRM-MLP from \citet{jolicoeurmartineau2025trm}.
Fair comparison therefore requires matching not only the test puzzles but also the inference budget (unroll steps and damping; for FPRM, also residual tolerance and step-size decay).
On Maze and $S_5$ our released runs use $1{,}024$ Picard steps at $\beta=0.3$ and up to $256$ at $\beta=0.7$, respectively (full protocol in \secref{sec:eval}).
The training recipes also differ (for example, GFPR uses dihedral augmentation on Maze-Hard and FPRM does not), so Table~\ref{tab:main} compares published systems and does not by itself isolate the effect of the state space.
Unless noted, every method decodes a single trajectory.

\subsection{Results}
\label{sec:exp-results}

\begin{table}[t]
  \centering
  \caption{Exact-match accuracy (\%). ``--'' was not reported.
  Quoted baselines are from \citet{movahedi2026fprm}, except TRM-MLP~\citep{jolicoeurmartineau2025trm}.
  $^{\dagger}$One trajectory.
  $^{\ddagger}$A sample of 23{,}680 Sudoku test puzzles.
  Slash-separated parameter counts correspond to Sudoku-Extreme, Maze-Hard, and $S_5$, in that order.}
  \label{tab:main}
  \small
  \begin{tabular}{lcccc}
    \toprule
    Model & Params & Sudoku-Extreme & Maze-Hard & $S_5$ \\
    \midrule
    HRM & 27M & 55.0 & 74.5 & -- \\
    TRM (attention) & 7M & 74.7 & 85.3 & 39.4 \\
    TRM (MLP) & 5M & 87.4 & -- & -- \\
    TRM + causal conv & 7M & -- & -- & 97.2 \\
    EqR & 7M & 93.0$^{\dagger}$ & -- & -- \\
    FPRM & 7M & 94.2 & 87.0 & 98.8 \\
    \midrule
    GFPR (ours) & 7M / 7M / 5.6M & \textbf{95.1}$^{\dagger}$ & \textbf{92.0} & \textbf{100.0} \\
    GFPR with constraints (ours) & 6.4M / 6.8M / 1.2M & \textbf{95.0}$^{\dagger\ddagger}$ & \textbf{96.0} & \textbf{100.0} \\
    \bottomrule
  \end{tabular}
\end{table}

Table~\ref{tab:main} summarizes the results.
On Sudoku-Extreme, GFPR solves 95.1\% of the 422{,}786 test puzzles (96.9\% of cells) from a single trajectory, above FPRM (94.2\%) and the other single-trajectory models at 5--7M parameters.
EqR reaches 99.8\% when it runs 128 restarts per puzzle and selects by residual~\citep{huang2026eqr}; this breadth scaling is orthogonal to GFPR and could be combined with it.
On Maze-Hard, GFPR solves 92.0\% of the 1{,}000 test mazes, compared with 87.0\% for FPRM.
FPRM trains on Maze-Hard without augmentation, whereas GFPR uses dihedral augmentation, as TRM does.
On $S_5$, trained on sequences of at most 32 updates and tested at length 128 with 320 effective layers, FPRM reaches 98.8\% and plain TRM 39.4\%; a causal convolution, which is not part of the original TRM, raises TRM to 97.2\% (means over seeds, $\pm0.9$, $\pm1.9$, $\pm2.5$;~\citealp{movahedi2026fprm}).
GFPR, a 5.6M product-of-simplices model and a single seed, reaches 100.0\% sequence accuracy on the 1{,}000 test sequences, which implies final-state accuracy.
The constrained models of \secref{sec:method-constraints} help most on Maze-Hard, where the unit-flow polytope raises exact match from 92.0\% to 96.0\%.
On $S_5$ both the softmax and the $\alpha$-entmax variants reach 100.0\%.
On Sudoku, with $\alpha=1.75$, the consistency polytope scores 94.99\% against 94.95\% for plain GFPR on the same 23{,}680 random test puzzles, so it matches but does not yet improve on the product of simplices.

\paragraph{What the belief state shows.}
Because the state is the prediction, its failures can be inspected without a decoder.
On Maze-Hard, 99.67\% of cells carry the correct parent pointer although only 92.0\% of mazes are solved exactly, and 99.7\% of settled parent fields trace a connected route from goal to start: most failures are a valid route that differs from the labeled one, not a broken field.
On Sudoku the gap is similar, with 96.9\% of cells correct against 95.1\% of grids.
The residual also serves as a stopping rule: on $S_5$ at length 128, stopping once the largest per-site total-variation step falls below $5\cdot10^{-3}$ takes 81 steps on average against a cap of 256, and every test sequence is still correct.

\subsection{Language-model results}
\label{sec:exp-lm}

\begin{table}[t]
  \centering
  \caption{Zero-shot multiple-choice accuracy (\%). ARC-Easy, SciQ, and PIQA are accuracy; HellaSwag is length-normalized accuracy.
  GFPR-LM is the 201M FineWeb-Edu run (EMA weights, $\beta=0.5$, 6B tokens), scored by next-token log-likelihood after $K$ damped Picard steps from the uniform state.
  GPT-2 small (124M) and medium (355M) use one forward pass in the same harness.
  All GFPR-LM columns are official full-split figures for this checkpoint (sweep through $K{=}32$ in \secref{sec:supp-lm-eval}); $K{=}24$ is the training depth, not a test-set choice.
  Bold marks the best of GPT-2 and $K{=}24$; the $K{=}8$ and $K{=}16$ columns show how the scores depend on depth.}
  \label{tab:lm}
  \small
  \begin{tabular}{lcccccc}
    \toprule
    & & & \multicolumn{4}{c}{GFPR-LM, Picard steps $K$} \\
    \cmidrule(l){4-7}
    Task & GPT-2 S & GPT-2 M & 1 & 8 & 16 & 24 \\
    \midrule
    ARC-Easy & 44.2 & 49.2 & 32.0 & 28.4 & 49.5 & \textbf{51.9} \\
    SciQ & 72.5 & \textbf{75.3} & 37.4 & 32.9 & 76.1 & 74.4 \\
    PIQA & 62.4 & \textbf{66.9} & 55.6 & 53.8 & 62.8 & 63.1 \\
    HellaSwag & 31.2 & \textbf{39.5} & 27.8 & 26.0 & 32.2 & 33.2 \\
    \bottomrule
  \end{tabular}
\end{table}

Table~\ref{tab:lm} compares the FineWeb-Edu model with GPT-2.
One Picard step is a weak language model on every task.
Twenty-four steps, the depth used in training, raise ARC-Easy from 32.0\% to 51.9\%, SciQ from 37.4\% to 74.4\%, PIQA from 55.6\% to 63.1\%, and HellaSwag from 27.8\% to 33.2\%.
The gain is not gradual: up to eight steps the scores stay at or below the one-step level, and most of the improvement arrives between $K{=}8$ and $K{=}16$.
Training settles for 16 steps before its differentiated tail, and the beliefs become useful at about that depth.
At $K=24$ GFPR-LM is above GPT-2 small on all four tasks and above GPT-2 medium on ARC-Easy; it remains below GPT-2 medium on SciQ, PIQA, and HellaSwag.
The comparison is not compute-matched: GPT-2 pays one trunk pass per token, while each Picard step is a full pass over the prompt and continuation.
The scores are likelihoods of answer choices, not samples from the wavefront decoder of \secref{sec:method-lm}.
Held-out FineWeb-Edu validation perplexity is 18.9.
The experiment is small by current standards: one 201M model, 6B tokens, and four likelihood-based benchmarks.
We read it as a proof of concept rather than a competitive language model: the same belief-state update, with the output simplex enlarged to the vocabulary, trains into a working language model whose quality comes from iteration (protocol in \secref{sec:supp-lm-eval}).

\section{Related Work}
\label{sec:related}

Recurrence in neural nets is older than the current puzzle models.
Hopfield networks already iterate a state that is also the output: a binary pattern is updated until it matches a stored memory~\citep{hopfield1982}.
Almeida--Pineda nets inject a constant input at every step and run until a fixed point~\citep{almeida1987,pineda1987}; written as $h^{l+1}=f(h^l,x;w)$ this is depth recurrence with $x$ clamped.
Classic graph neural nets do the same on a graph, but they constrain the transition to be a contraction so the fixed point is unique, and they still decode with a separate head~\citep{scarselli2009gnn}.

Weight-tied depth then reappeared as a way to spend test-time compute without growing the parameter count.
The Universal Transformer ties one block across depth and can stop with ACT~\citep{dehghani2019ut,graves2016act}.
Looped Transformers put a tied core between an untied prelude and coda, re-inject the input each step, and raise the loop count at test time~\citep{giannou2023looped,geiping2025recurrent}.
HRM nests two timescales and TRM collapses them to one stream; both can stop with ACT or a step cap~\citep{wang2025hrm,jolicoeurmartineau2025trm}.
The iterate in these models is a free latent, and a head turns the last state into symbols.

Deep equilibrium models treat an infinitely deep weight-tied stack as the root of $z=f_\theta(z;x)$ and differentiate at that root~\citep{bai2019deq}.
FPRM and EqR use the same idea as the reasoning procedure itself~\citep{movahedi2026fprm,huang2026eqr}.
FPRM damps the Picard step and stops when the latent residual is small.
EqR treats the map as an attractor landscape and can spend extra compute on restarts.
In both, the state remains a vector in $\mathbb{R}^d$; a fixed point is not guaranteed by the architecture.

GFPR is an Almeida--Pineda-style depth recurrence: the input is injected every step, and the state is a field of categorical beliefs.
As in a Hopfield network, the state is itself the output, but it is a learned distribution per site rather than a binary pattern matched to stored memories.
The same state can carry structural constraints on a task (\secref{sec:method-constraints}), trained with a Fenchel--Young loss~\citep{blondel2020fenchel}.

On language, a standard Transformer has no depth recurrence: each block has its own weights and is visited once~\citep{radford2019language}.
Chain-of-thought spends compute by writing tokens~\citep{wei2022cot,kojima2022zeroshot}.
Latent looped language models iterate a hidden state~\citep{geiping2025recurrent}.
GFPR iterates a belief over the vocabulary at each site (\secref{sec:method-lm}, \secref{sec:exp-lm}).

\section{Conclusion}
\label{sec:conclusion}

We argued that a looped reasoner should iterate its prediction rather than a latent code decoded at the end.
In GFPR every intermediate state is a field of categorical beliefs that can be read as an answer, its residual measures how much that answer moves, and task structure can be imposed on it as a convex set.
A fixed point then exists for any parameters, as a consequence of the design rather than its purpose; which one is reached is decided by training.

On Sudoku-Extreme, Maze-Hard, and $S_5$, this state is enough to exceed the published single-trajectory numbers of latent looped reasoners at similar scale, and on the puzzle of \Figref{fig:fp-residual-compare} GFPR settles to a fixed point where FPRM's halt is not one.
Adding task-specific convex structure helps where the relaxation is informative: the unit-flow readout raises Maze-Hard from 92.0\% to 96.0\%, while the Sudoku consistency readout so far only matches the product-of-simplices baseline.
The language-model experiment is small, but it shows that the same update works when each site is a distribution over a vocabulary: after 24 Picard steps a 201M model trained on 6B tokens is competitive with GPT-2 small and medium on zero-shot multiple choice, and that quality appears only after iteration.

The evidence has clear limits.
The baselines are quoted rather than re-run under a matched training recipe and inference budget, so the share of the gain due to the state space is not isolated.
The language-model comparison is not compute-matched and scores likelihoods rather than generated text.
A few Sudoku puzzles settle to a stable but wrong grid; because the state can be checked against the constraints without labels, restarts selected by constraint violation are a natural next step.

\clearpage
\bibliographystyle{iclr2027_conference}
\bibliography{refs}

@inproceedings{wei2022cot,
  title={Chain-of-Thought Prompting Elicits Reasoning in Large Language Models},
  author={Wei, Jason and Wang, Xuezhi and Schuurmans, Dale and Bosma, Maarten and Ichter, Brian and Xia, Fei and Chi, Ed H. and Le, Quoc V. and Zhou, Denny},
  booktitle={Advances in Neural Information Processing Systems},
  volume={35},
  year={2022}
}

@inproceedings{kojima2022zeroshot,
  title={Large Language Models are Zero-Shot Reasoners},
  author={Kojima, Takeshi and Gu, Shixiang Shane and Reid, Machel and Matsuo, Yutaka and Iwasawa, Yusuke},
  booktitle={Advances in Neural Information Processing Systems},
  volume={35},
  pages={22199--22213},
  year={2022}
}

@misc{snell2024testtime,
  title={Scaling {LLM} Test-Time Compute Optimally can be More Effective than Scaling Model Parameters},
  author={Snell, Charlie and Lee, Jaehun and Xu, Kelvin and Kumar, Aviral},
  year={2024},
  eprint={2408.03314},
  archivePrefix={arXiv},
  primaryClass={cs.LG}
}

@misc{movahedi2026fprm,
  title={Fixed-Point Reasoners: Stable and Adaptive Deep Looped Transformers},
  author={Sajad Movahedi and Vera Milovanovi{\c{c}} and Shlomo Libo Feigin and Alexander Theus and Thomas Hofmann and Valentina Boeva and T. Konstantin Rusch and Antonio Orvieto},
  year={2026},
  eprint={2606.18206},
  archivePrefix={arXiv},
  primaryClass={cs.AI}
}

@misc{wang2025hrm,
  title={Hierarchical Reasoning Model},
  author={Guan Wang and Jin Li and Yuhao Sun and Xing Chen and Changling Liu and Yue Wu and Meng Lu and Sen Song and Yasin Abbasi Yadkori},
  year={2025},
  eprint={2506.21734},
  archivePrefix={arXiv},
  primaryClass={cs.AI}
}

@misc{jolicoeurmartineau2025trm,
  title={Less is More: Recursive Reasoning with Tiny Networks},
  author={Alexia Jolicoeur-Martineau},
  year={2025},
  eprint={2510.04871},
  archivePrefix={arXiv},
  primaryClass={cs.LG}
}

@misc{huang2026eqr,
  title={Equilibrium Reasoners: Learning Attractors Enables Scalable Reasoning},
  author={Benhao Huang and Zhengyang Geng and Zico Kolter},
  year={2026},
  eprint={2605.21488},
  archivePrefix={arXiv},
  primaryClass={cs.LG}
}

@inproceedings{bai2019deq,
  title={Deep Equilibrium Models},
  author={Shaojie Bai and J. Zico Kolter and Vladlen Koltun},
  booktitle={Advances in Neural Information Processing Systems},
  volume={32},
  year={2019}
}

@inproceedings{kerg2019nnrnn,
  title={Non-normal Recurrent Neural Network ({nnRNN}): learning long time dependencies while improving expressivity with transient dynamics},
  author={Giancarlo Kerg and Kyle Goyette and Maximilian Puelma Touzel and Gauthier Gidel and Eugene Vorontsov and Yoshua Bengio and Guillaume Lajoie},
  booktitle={Advances in Neural Information Processing Systems},
  volume={32},
  year={2019}
}

@inproceedings{dehghani2019ut,
  title={Universal Transformers},
  author={Mostafa Dehghani and Stephan Gouws and Oriol Vinyals and Jakob Uszkoreit and Łukasz Kaiser},
  booktitle={International Conference on Learning Representations},
  year={2019}
}

@article{graves2016act,
  title={Adaptive Computation Time for Recurrent Neural Networks},
  author={Alex Graves},
  journal={arXiv preprint arXiv:1603.08983},
  year={2016}
}

@article{brouwer1911,
  title={{\"U}ber Abbildung von Mannigfaltigkeiten},
  author={Brouwer, L. E. J.},
  journal={Mathematische Annalen},
  volume={71},
  pages={97--115},
  year={1911}
}

@article{banach1922,
  title={Sur les op{\'e}rations dans les ensembles abstraits et leur application aux {\'e}quations int{\'e}grales},
  author={Banach, Stefan},
  journal={Fundamenta Mathematicae},
  volume={3},
  number={1},
  pages={133--181},
  year={1922},
  doi={10.4064/fm-3-1-133-181}
}

@book{ortega2000nonlinear,
  title={Iterative Solution of Nonlinear Equations in Several Variables},
  author={Ortega, James M. and Rheinboldt, Werner C.},
  publisher={SIAM},
  series={Classics in Applied Mathematics},
  volume={30},
  year={2000},
  note={Originally published 1970},
  doi={10.1137/1.9780898719468}
}

@book{kelley1995iterative,
  title={Iterative Methods for Linear and Nonlinear Equations},
  author={Kelley, C. T.},
  publisher={SIAM},
  series={Frontiers in Applied Mathematics},
  volume={16},
  year={1995},
  doi={10.1137/1.9781611970944}
}

@book{horn2012matrix,
  title={Matrix Analysis},
  author={Horn, Roger A. and Johnson, Charles R.},
  edition={2nd},
  publisher={Cambridge University Press},
  year={2012}
}

@book{trefethen2005spectra,
  title={Spectra and Pseudospectra: The Behavior of Nonnormal Matrices and Operators},
  author={Trefethen, Lloyd N. and Embree, Mark},
  publisher={Princeton University Press},
  year={2005}
}

@inproceedings{merrill2024illusion,
  title={The Illusion of State in State-Space Models},
  author={Merrill, William and Petty, Jackson and Sabharwal, Ashish},
  booktitle={Proceedings of the 41st International Conference on Machine Learning},
  series={Proceedings of Machine Learning Research},
  volume={235},
  pages={35492--35506},
  publisher={PMLR},
  year={2024}
}

@article{blondel2020fenchel,
  title={Learning with {Fenchel-Young} Losses},
  author={Blondel, Mathieu and Martins, Andr{\'e} F. T. and Niculae, Vlad},
  journal={Journal of Machine Learning Research},
  volume={21},
  number={35},
  pages={1--69},
  year={2020}
}

@techreport{radford2019language,
  title={Language Models are Unsupervised Multitask Learners},
  author={Radford, Alec and Wu, Jeffrey and Child, Rewon and Luan, David and Amodei, Dario and Sutskever, Ilya},
  institution={OpenAI},
  year={2019}
}

@inproceedings{penedo2024fineweb,
  title={The {FineWeb} Datasets: Decanting the Web for the Finest Text Data at Scale},
  author={Penedo, Guilherme and Kydl{\'\i}{\v{c}}ek, Hynek and Ben Allal, Loubna and Lozhkov, Anton and Mitchell, Margaret and Raffel, Colin and Von Werra, Leandro and Wolf, Thomas},
  booktitle={Advances in Neural Information Processing Systems},
  volume={37},
  pages={30811--30849},
  year={2024}
}

@article{clark2018arc,
  title={Think you have Solved Question Answering? Try {ARC}, the {AI2} Reasoning Challenge},
  author={Clark, Peter and Cowhey, Isaac and Etzioni, Oren and Khot, Tushar and Sabharwal, Ashish and Schoenick, Carissa and Tafjord, Oyvind},
  journal={arXiv preprint arXiv:1803.05457},
  year={2018}
}

@inproceedings{welbl2017sciq,
  title={Crowdsourcing Multiple Choice Science Questions},
  author={Welbl, Johannes and Liu, Nelson F. and Gardner, Matt},
  booktitle={Proceedings of the 3rd Workshop on Noisy User-generated Text},
  pages={94--106},
  year={2017},
  address={Copenhagen, Denmark},
  publisher={Association for Computational Linguistics}
}

@inproceedings{bisk2020piqa,
  title={{PIQA}: Reasoning about Physical Commonsense in Natural Language},
  author={Bisk, Yonatan and Zellers, Rowan and Gao, Jianfeng and Choi, Yejin},
  booktitle={Proceedings of the AAAI Conference on Artificial Intelligence},
  volume={34},
  pages={7432--7439},
  year={2020}
}

@inproceedings{peters2019sparse,
  title={Sparse Sequence-to-Sequence Models},
  author={Peters, Ben and Niculae, Vlad and Martins, Andr{\'e} F. T.},
  booktitle={Proceedings of the 57th Annual Meeting of the Association for Computational Linguistics},
  year={2019},
  pages={1504--1519},
  publisher={Association for Computational Linguistics},
  doi={10.18653/v1/P19-1146},
  url={https://arxiv.org/abs/1905.05702}
}

@inproceedings{zellers2019hellaswag,
  title={{HellaSwag}: Can a Machine Really Finish Your Sentence?},
  author={Zellers, Rowan and Holtzman, Ari and Bisk, Yonatan and Farhadi, Ali and Choi, Yejin},
  booktitle={Proceedings of the 57th Annual Meeting of the Association for Computational Linguistics},
  year={2019}
}

@article{hopfield1982,
  title={Neural networks and physical systems with emergent collective computational abilities},
  author={Hopfield, J. J.},
  journal={Proceedings of the National Academy of Sciences},
  volume={79},
  number={8},
  pages={2554--2558},
  year={1982},
  doi={10.1073/pnas.79.8.2554}
}

@inproceedings{almeida1987,
  title={A learning rule for asynchronous perceptrons with feedback in a combinatorial environment},
  author={Almeida, Lu{\'\i}s B.},
  booktitle={Proceedings of the IEEE First International Conference on Neural Networks},
  volume={2},
  pages={609--618},
  year={1987}
}

@article{pineda1987,
  title={Generalization of back-propagation to recurrent neural networks},
  author={Pineda, Fernando J.},
  journal={Physical Review Letters},
  volume={59},
  number={19},
  pages={2229--2232},
  year={1987},
  doi={10.1103/PhysRevLett.59.2229}
}

@article{scarselli2009gnn,
  title={The Graph Neural Network Model},
  author={Scarselli, Franco and Gori, Marco and Tsoi, Ah Chung and Hagenbuchner, Markus and Monfardini, Gabriele},
  journal={IEEE Transactions on Neural Networks},
  volume={20},
  number={1},
  pages={61--80},
  year={2009},
  doi={10.1109/TNN.2008.2005605}
}

@inproceedings{giannou2023looped,
  title={Looped Transformers as Programmable Computers},
  author={Giannou, Angeliki and Rajput, Shashank and Sohn, Jy-yong and Lee, Kangwook and Lee, Jason D. and Papailiopoulos, Dimitris},
  booktitle={International Conference on Machine Learning},
  pages={11398--11442},
  year={2023}
}

@inproceedings{geiping2025recurrent,
  title={Scaling up Test-Time Compute with Latent Reasoning: A Recurrent Depth Approach},
  author={Geiping, Jonas and McLeish, Sean and Jain, Neel and Kirchenbauer, John and Singh, Siddharth and Bartoldson, Brian R. and Kailkhura, Bhavya and Bhatele, Abhinav and Goldstein, Tom},
  booktitle={Advances in Neural Information Processing Systems},
  volume={38},
  pages={41340--41391},
  year={2025}
}

\clearpage
\appendix
\section{FPRM and GFPR side by side}
\label{sec:compare}

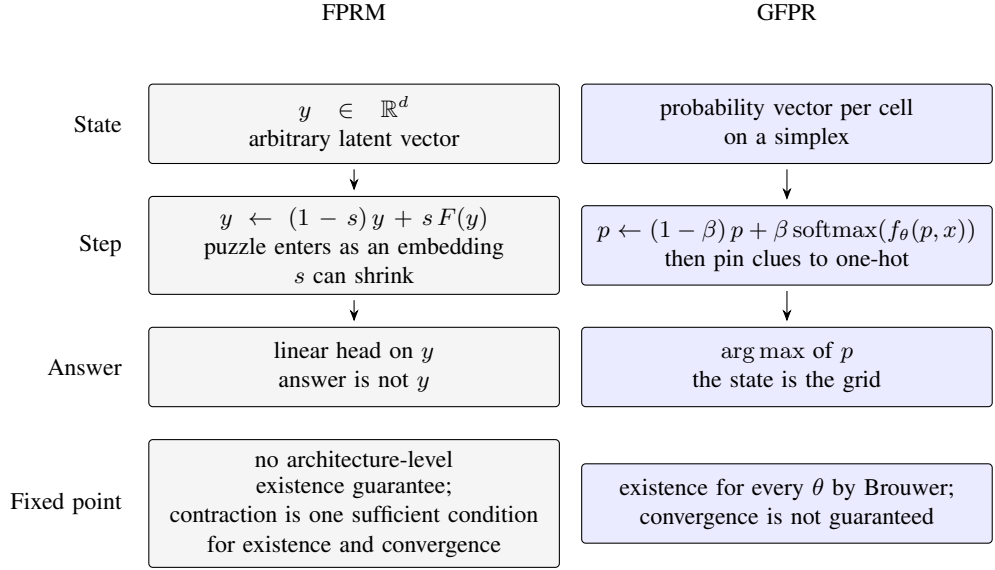
\begin{figure}[H]
\centering
\begin{tikzpicture}[
  font=\footnotesize,
  arr/.style={-{Stealth[length=1.6mm]}, shorten >=2pt, shorten <=2pt},
  box/.style={draw, rounded corners=1.5pt, align=center, inner sep=4pt,
              minimum height=1.05cm, anchor=center},
  lab/.style={draw=none, align=right, font=\footnotesize, inner sep=2pt, anchor=center},
  hdr/.style={draw=none, font=\small, align=center, inner sep=1pt, anchor=center},
]
\matrix (m) [
  matrix of nodes,
  column sep=8pt,
  row sep=12pt,
  nodes={box, text width=5.15cm},
  column 1/.style={nodes={lab, text width=1.55cm}},
  row 1/.style={nodes={hdr, text width=5.15cm}},
  row 5/.style={row sep=16pt},
] {
& FPRM & GFPR \\
State &
|[fill=black!4]| {$y\in\mathbb{R}^{d}$\\[1pt] arbitrary latent vector} &
|[fill=blue!8]| {probability vector per cell\\[1pt] on a simplex} \\
Step &
|[fill=black!4]| {$y \leftarrow (1-s)\,y + s\,F(y)$\\[1pt] puzzle enters as an embedding\\[1pt] $s$ can shrink} &
|[fill=blue!8]| {$p \leftarrow (1-\beta)\,p + \beta\,\softmax(f_\theta(p,x))$\\[1pt] then pin clues to one-hot} \\
Answer &
|[fill=black!4]| {linear head on $y$\\[1pt] answer is not $y$} &
|[fill=blue!8]| {$\arg\max$ of $p$\\[1pt] the state is the grid} \\
Fixed point &
|[fill=black!4]| {no architecture-level existence guarantee;\\[1pt] contraction is one sufficient condition\\[1pt] for existence and convergence} &
|[fill=blue!8]| {existence for every $\theta$ by Brouwer;\\[1pt] convergence is not guaranteed} \\
};
\draw[arr] (m-2-2) -- (m-3-2);
\draw[arr] (m-3-2) -- (m-4-2);
\draw[arr] (m-2-3) -- (m-3-3);
\draw[arr] (m-3-3) -- (m-4-3);
\end{tikzpicture}
\caption{One test-time step.
Arrows are the loop.
The bottom row is a property of that loop, not another update.
FPRM reads the grid from a head on an arbitrary latent.
GFPR reads the grid from the belief it iterates, after pinning the given clues.}
\label{fig:fprm-gfpr}
\end{figure}

Figure~\ref{fig:fprm-gfpr} is one step of each method.
The first three rows are the loop.
The last row is not a further step.
It is the reason a fixed point does or does not come for free.

FPRM can stop when the step is small; GFPR runs for a fixed step budget at test time.
They do not certify the same thing.
In FPRM the step is motion of $y$.
A head still turns $y$ into digits; contraction is sufficient for a fixed point of $F$ and for local convergence, but neither is guaranteed by the architecture.
In GFPR the step is motion of $p$, and $p$ is already the prediction.
Clues stay pinned, the state stays in a compact product of simplices, so Brouwer gives a fixed point for every $\theta$; which equilibrium the iteration reaches is still a training question.

\section{Model and training settings}
\label{sec:hparams}

Table~\ref{tab:hparams} lists the settings of the two released GFPR runs.
Both use the same Transformer trunk over all sites.
Inside one application of $F_\theta$ the trunk is run for a few self-conditioning passes: each pass takes the previous pass's softmax output as extra input, and the last one is the output of $F_\theta$.
Before each softmax the logits are soft-capped as $15\tanh(\ell/15)$.

The training rollout (Section~3.2 of the main text) has two phases.
The first runs without gradients for at most $D$ steps (random on Sudoku, fixed on Maze) and stops early once every puzzle in the batch has kept its maximum per-site TV step below a tolerance for two consecutive steps.
On Sudoku, $D$ is drawn per batch as $D=1+\mathrm{Poisson}(e^{\tau})$ with $\tau\sim\mathcal{N}(\log\bar D-\sigma^2/2,\ \sigma^2)$, which has mean close to $\bar D$ and a heavy right tail; on Maze-Hard, $D$ is fixed.
The second phase differentiates a tail of $m$ steps, and the loss is applied to its final state.
With probability $0.25$ a rollout starts from a random Dirichlet state instead of the uniform one.
Evaluation and checkpoint selection use an exponential moving average of the weights.

\begin{table}[H]
  \centering
  \caption{GFPR settings for the released runs. $K$ is the number of output symbols and $A$ the number of auxiliary coordinates; for S5, $K=121$ counts the 120 permutations and the pad token. Rollout depth is the no-gradient prefix: ``$\bar D{=}32$'' is the log-normal Poisson draw in the main text, and the TV tolerance is its early-stop threshold. ``cosine@150k'' starts the learning-rate decay at step 150{,}000. S5 applies a causal depthwise convolution of kernel size 4 to the state before adding operation embeddings; the other runs do not.}
  \label{tab:hparams}
  \footnotesize
  \setlength{\tabcolsep}{4pt}
  \begin{tabular}{@{}lccc@{}}
    \toprule
    & Sudoku-Extreme & Maze-Hard & S5 \\
    \midrule
    Parameters & 7.16M & 6.81M & 5.60M \\
    Width / layers / heads & 256 / 9 / 8 & 512 / 2 / 8 & 224 / 9 / 8 \\
    $K$ / $A$ & 9 / 16 & 6 / 8 & 121 / 16 \\
    Self-conditioning passes & 4 & 2 & 4 \\
    Damping $\beta$ & 0.7 & 0.3 & 0.7 \\
    Rollout depth & $\bar D{=}32$ & 40 & $\bar D{=}32$ \\
    Early-stop tolerance (TV) & $5\cdot10^{-3}$ & $10^{-5}$ & $5\cdot10^{-3}$ \\
    Differentiated tail & 8 & 10 & 8 \\
    $\lambda_{\mathrm{aux}}$ / $\lambda_{\mathrm{res}}$ & 0.02 / 0.02 & 0.02 / 0 & 0.02 / 0.02 \\
    Batch size & 128 & 8 & 128 \\
    Learning rate & $3\cdot10^{-4}$ & $3\cdot10^{-4}$ & $3\cdot10^{-4}$ \\
    LR schedule & cosine@150k & cosine & cosine@150k \\
    Warmup / steps & 2k / 200k & 1k / 60k & 2k / 200k \\
    Selected checkpoint & 54k & 42k & 12k \\
    Dropout / EMA decay & 0.1 / 0.9999 & 0.1 / 0.9999 & 0.1 / 0.9999 \\
    Augmentation & Sudoku symmetries & dihedral & none \\
    \bottomrule
  \end{tabular}
\end{table}

\section{Evaluation protocol}
\label{sec:eval}

\paragraph{Sudoku-Extreme.}
Checkpoints are selected on a stratified 1{,}000-puzzle validation sample of the \texttt{test\_hard} split of \texttt{sapientinc/sudoku-extreme-1k}; the remaining \texttt{test\_hard} puzzles form a disjoint local test split for development.
The selected checkpoint reaches 94.0\% exact match on a 500-puzzle subsample of the validation slice with a 20{,}000-step budget.
Headline evaluation is computed on all 422{,}786 puzzles of \texttt{sapientinc/sudoku-extreme} (\texttt{test.csv})---a separate Hugging Face release from \texttt{sudoku-extreme-1k}---with the same cap as the released FPRM Sudoku-Extreme evaluation: a single trajectory from the uniform state, $\beta=0.7$, and at most 35{,}000 damped Picard steps (early stop when every site's total-variation step falls below $5\cdot10^{-3}$ for two consecutive steps); that file is never used for checkpoint selection.
Exact match is 95.11\% and cell accuracy 96.93\%.

\paragraph{Maze-Hard.}
Evaluation uses all 1{,}000 test mazes of \texttt{sapientinc/maze-30x30-hard-1k}, exactly 1{,}024 damped Picard steps, and $\beta=0.3$.
Exact match is 92.0\%, cell accuracy 99.67\%, and 99.7\% of settled parent fields trace a connected route from goal to start.

\paragraph{S5.}
Training uses 2{,}000{,}000 sequences of 32 group updates, and evaluation uses all 1{,}000 test sequences of length 128.
Inference runs at most 256 damped Picard steps at $\beta=0.7$ from the uniform state and stops early when the maximum per-site total variation falls below $5\cdot10^{-3}$.
On the selected checkpoint every test sequence is correct at every prefix length, so both sequence accuracy and final-state accuracy are 100\%.
The mean number of steps to that tolerance is 81.

\paragraph{Parent-field target.}
A breadth-first tree defines a unique parent only once the order in which neighbors are expanded is fixed, and most Maze-Hard instances have several shortest routes.
We recover the expansion order of the dataset generator from the data and check that it reproduces every labeled route, so the traced route is compared with the same route that the published numbers score.

\section{Task-specific convex sets}
\label{sec:supp-constraints}

This section gives the convex sets used by the constrained models in \secref{sec:method-constraints}.

\paragraph{Sudoku.}
For the Sudoku-Extreme dataset, let $q_{rcd}$ denote the coordinate associated with digit $d\in\{1,\ldots,9\}$ at cell $(r,c)$, and let $\mathcal{B}$ denote the collection of nine $3\times3$ boxes. For a puzzle $x$, let $d_{rc}$ denote the given digit at each clue cell $(r,c)$. We define
\begin{equation}
\label{eq:sudoku-polytope}
\mathcal{C}_{\mathrm{Sud}}(x)
=
\left\{
q\in[0,1]^{9\times9\times9}:
\begin{array}{ll}
  \displaystyle\sum_{d=1}^9 q_{rcd}=1
    & \forall r,c,\\[2pt]
  \displaystyle\sum_{c=1}^9 q_{rcd}=1
    & \forall r,d,\\[2pt]
  \displaystyle\sum_{r=1}^9 q_{rcd}=1
    & \forall c,d,\\[2pt]
  \displaystyle\sum_{(r,c)\in B} q_{rcd}=1
    & \forall B\in\mathcal{B},d,\\[2pt]
  q_{rc,d_{rc}}=1
    & \text{for every clue cell }(r,c).
\end{array}
\right\}.
\end{equation}
The first family of equalities makes $(q_{rc1},\ldots,q_{rc9})$ a probability vector for every cell. The remaining equalities require each digit to have total coordinate sum one in every row, column, and box, while the final constraints fix the clue coordinates. Every valid completed Sudoku grid corresponds to an integral point of $\mathcal{C}_{\mathrm{Sud}}(x)$. The converse does not hold for arbitrary feasible points because the polytope can contain fractional solutions; consequently, feasibility in $\mathcal{C}_{\mathrm{Sud}}(x)$ does not by itself imply a valid discrete Sudoku completion.

We use the power regularizer $\Omega_\alpha$ defined in \secref{sec:method-constraints} and define
\begin{equation}
\label{eq:sudoku-primal}
  q^*(z)
  =
  \operatorname*{arg\,max}_{q\in\mathcal{C}_{\mathrm{Sud}}(x)}
  \left\{
    z^\top q-\Omega_\alpha(q)
  \right\}.
\end{equation}
We evaluate Eq.~\eqref{eq:sudoku-primal} through its convex dual. After fixing the clue coordinates, let $F$ denote the remaining coordinates and write the equality constraints as
$A_Fq_F=\widetilde{b}_x$. For a dual variable $\lambda$, define
$s=z_F-A_F^\top\lambda$. Maximization of the Lagrangian with respect to the primal coordinates yields
\begin{equation}
\label{eq:sudoku-dual-map}
  q_i(\lambda)
  =
  \min\left\{
    1,
    \bigl[(\alpha-1)[s_i]_+\bigr]^{1/(\alpha-1)}
  \right\}.
\end{equation}
The dual problem is solved numerically, and the resulting primal point is recovered from Eq.~\eqref{eq:sudoku-dual-map}. Numerical feasibility is measured by the equality residual
$\|A_Fq_F(\lambda)-\widetilde{b}_x\|$.

\begin{figure*}[t]
  \centering
  \begin{minipage}[t]{0.52\textwidth}
    \vspace{0pt}
    \centering
    \includegraphics[width=\linewidth]{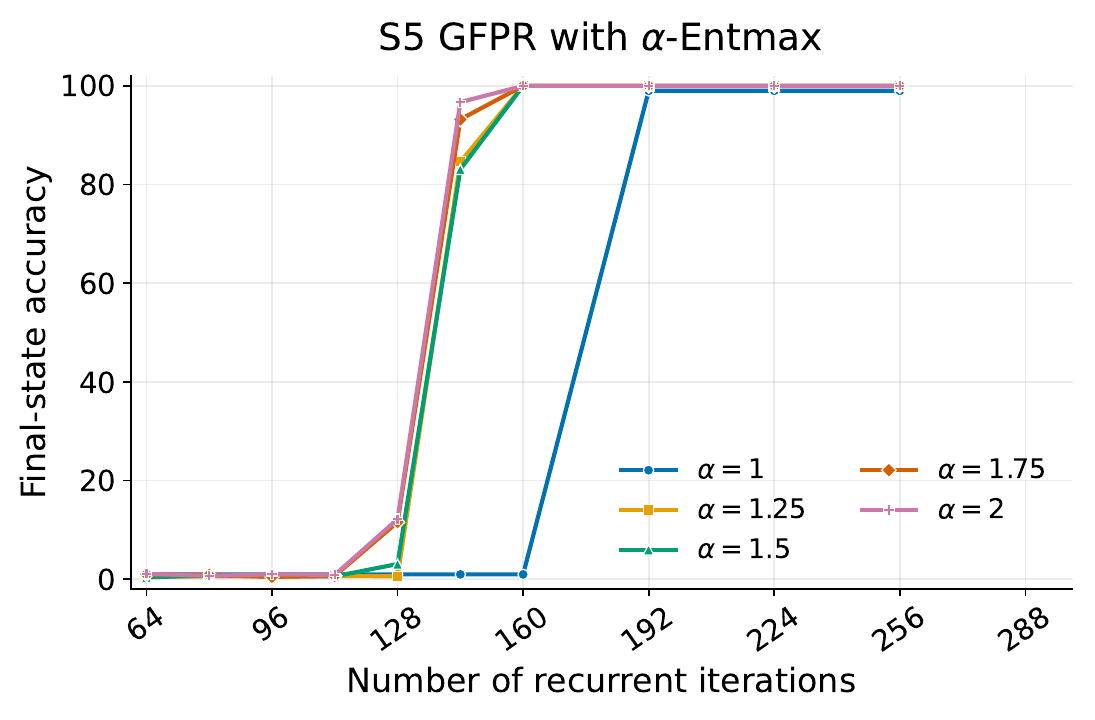}
    \captionof{figure}{S5 final-state accuracy versus the number of recurrent iterations for different Entmax orders $\alpha$.}
    \label{fig:s5-alpha-ablation}
  \end{minipage}\hfill
  \begin{minipage}[t]{0.46\textwidth}
    \vspace{0pt}
    \centering
    \small
    \begin{tabular}{c|rrrrrr}
      \toprule
      \multirow{2}{*}{$\alpha$}
        & \multicolumn{6}{c}{Number of recurrent iterations} \\
      \cmidrule(l){2-7}
        & 8 & 16 & 32 & 64 & 128 & 160 \\
      \midrule
      1.00 & 49.0 & 62.1 & 71.9 & 77.4 & 82.1 & 83.2 \\
      1.25 & 49.8 & 62.5 & 72.0 & 77.4 & 82.1 & 83.2 \\
      1.50 & 50.5 & 62.7 & 72.2 & 77.4 & 82.2 & 83.2 \\
      1.75 & 51.2 & 62.7 & 72.2 & 77.5 & 82.2 & 83.2 \\
      2.00 & 51.5 & 63.2 & 72.2 & 77.6 & 82.2 & 83.2 \\
      \bottomrule
    \end{tabular}
    \captionof{table}{Sudoku-Extreme exact-grid accuracy (\%) for different Entmax orders $\alpha$ and recurrent iteration budgets.}
    \label{tab:sudoku-alpha-ablation}
  \end{minipage}
\end{figure*}

The choice of $\alpha$ has only a modest effect on the number of recurrent iterations needed to reach the final accuracy: it shifts the transition point slightly, while the curves nearly coincide after sufficient iteration.
The S5 and Sudoku ablations are reported in \figref{fig:s5-alpha-ablation} and Table~\ref{tab:sudoku-alpha-ablation}.

\paragraph{Maze.}
For Maze-Hard dataset, let $G_x=(V_x,E_x)$ be the directed graph whose
vertices are the open cells and whose edge set contains both
orientations of every adjacent pair of open cells. Let
$B_x\in\mathbb{R}^{|V_x|\times|E_x|}$ be its node--edge incidence
matrix, with coefficient $+1$ for an outgoing edge and $-1$ for an
incoming edge.
To match the parent-pointer decoding direction, we orient the flow from the goal $g$ to the start $s$ and define $b_x\in\mathbb{R}^{|V_x|}$ by
\[
  (b_x)_v
  =
  \begin{cases}
    1, & v=g,\\
    -1, & v=s,\\
    0, & \text{otherwise}.
  \end{cases}
\]
The flow polytope $\mathcal{C}_{\mathrm{flow}}(x)$ is defined in \eqref{eq:maze-flow-polytope}.
The equality $B_xf=b_x$ imposes flow conservation at every vertex other than $g$ and $s$, together with net outflow one at $g$ and net inflow one at $s$. The continuous relaxation may contain fractional flows and nonzero circulations. Therefore, membership in $\mathcal{C}_{\mathrm{flow}}(x)$ does not imply that $f$ is the incidence vector of a simple path, nor does it imply shortestness or uniqueness. These properties are evaluated on the decoded discrete path separately.

\paragraph{Fenchel--Young training.}
Training uses the Fenchel--Young loss \eqref{eq:fy-structured} and its gradient $\nabla_z L_{\Omega,\mathcal{C}}(z,y)=\Psi_{\mathcal{C}(x)}(z)-y$ when the argmax in \eqref{eq:general-convex-step} is unique (Danskin's theorem).
The structured map can be applied either at every recurrent step or only at the terminal readout (\secref{sec:method-constraints} lists which construction each benchmark uses).
In the recurrent construction, the structured component of every iterate belongs to $\mathcal{C}(x)$, and differentiation through an unrolled recurrence includes derivatives of the recurrent applications of $\Psi_{\mathcal{C}(x)}$.
In the terminal-readout construction, the recurrence remains in the product-simplex state space of \secref{sec:method-state}, and $\Psi_{\mathcal{C}(x)}$ is applied only to the final score vector, so the gradient with respect to the final scores does not differentiate through the numerical iterations used to compute the structured prediction.

\section{Simplex language-model decode}
\label{sec:supp-lm-decode}

\Secref{sec:method-lm} is implemented for a byte-pair TinyStories model in \texttt{talgat/lm/}.
The trunk is a causal Transformer with weights tied across Picard steps.
Training is the relax-plus-tail unroll of \secref{sec:method-training}: a non-differentiated settle, then a differentiated tail, with next-token cross-entropy on the output block of $p_t$ against $x_{t+1}$.
Open-ended generation uses wavefront decode (\texttt{generate\_progressive} in \texttt{talgat/lm/code/decode\_lm.py}).
The defaults are soft-ahead width $W=8$, damping $\beta=0.5$, sharpening temperature $\tau=1/16$, and a commit when either the top-two margin or a total-variation stability test fires, after a minimum number of Picard steps since the previous commit.
There is no key--value cache: every Picard step is a full forward over the active length, so iteration counts describe this decoder rather than a cached autoregressive Transformer.
The FineWeb-Edu multiple-choice scores in \secref{sec:exp-lm} use fixed-$K$ teacher-forced unrolling, not wavefront sampling.

\section{FineWeb-Edu language-model evaluation}
\label{sec:supp-lm-eval}

The 201M run in \secref{sec:exp-lm} is trained on 6{,}000{,}148{,}480 FineWeb-Edu tokens for 36{,}622 steps.
The trunk has width 2048, 16 heads, and two weight-tied blocks; the vocabulary has 16{,}384 byte-pair tokens and 16 auxiliary coordinates; the context is 512.
Training uses $\beta=0.5$, 16 no-gradient steps and an 8-step differentiated tail, and Muon with AdamW.
Reported scores use EMA weights.

Zero-shot scoring is next-token log-likelihood of each answer choice, with a BOS token, full continuation, and left truncation only on overflow.
The state starts uniform; the iteration count $K$ is fixed (no residual halt).
Headline metrics are accuracy on ARC-Easy (test, $N{=}2376$), SciQ (test, $N{=}1000$), and PIQA (validation, $N{=}1838$), and length-normalized accuracy on HellaSwag (validation, $N{=}10042$).
Final arithmetic is FP32 states and activations with TF32 matmul.

Table~\ref{tab:lm-depth} is the official depth sweep for this checkpoint.
Most of the gain appears between $K{=}8$ and $K{=}24$.
The step $K{=}24\to 32$ does not improve all four tasks, which is why the main text uses the training depth $K{=}24$.
These scores do not show that the simplex state has reached a fixed point.

\begin{table}[H]
  \centering
  \caption{GFPR-LM 201M zero-shot accuracy (\%) versus Picard depth $K$. HellaSwag is length-normalized accuracy.}
  \label{tab:lm-depth}
  \small
  \begin{tabular}{lccccccc}
    \toprule
    Task & $K{=}1$ & 2 & 4 & 8 & 16 & 24 & 32 \\
    \midrule
    ARC-Easy & 32.0 & 32.1 & 29.4 & 28.4 & 49.5 & 51.9 & 51.9 \\
    SciQ & 37.4 & 34.6 & 29.9 & 32.9 & 76.1 & 74.4 & 74.3 \\
    PIQA & 55.6 & 55.6 & 54.5 & 53.8 & 62.8 & 63.1 & 63.2 \\
    HellaSwag & 27.8 & 27.6 & 26.8 & 26.0 & 32.2 & 33.2 & 33.1 \\
    \bottomrule
  \end{tabular}
\end{table}

GPT-2 small and medium in Table~\ref{tab:lm} of the main text use one forward pass, batch 16, bfloat16, in the same task configurations.
GPT-2 large (774M) on the same four tasks is 53.0 / 79.7 / 70.1 / 45.3; we omit it from the main table because it is several times larger than the 201M GFPR-LM run.

\section{FPRM diagnostic protocol}
\label{sec:fprm-protocol}

The traces in the residual-comparison figure of Section~2 of the main text use FPRM's public Sudoku checkpoint and its evaluation loop, run in float64 for 80 steps on one test puzzle, with step-size decay $0.997$ and patience $10$ as in the released configuration.
GFPR is run on the same puzzle with $\beta=0.7$ from the uniform state.
The residual is the RMS ratio $\|F(u)-u\|/\|u\|$ at every step.

\paragraph{Local amplification probe (power iteration).}
At state $u$ we set $J_F=\partial F/\partial u$ and access it only through Jacobian--vector products from reverse-mode autodiff on $F$.
With a unit probe $v_0$, for $t=1,\ldots,K_{\mathrm{pi}}$ we compute $w_t=J_F v_{t-1}$, normalize $v_t=w_t/\|w_t\|_2$, and report the probe value $\widehat{\alpha}(u)=\|w_{K_{\mathrm{pi}}}\|_2$.
For a diagonalizable $J_F$ with a simple dominant eigenvalue, $\widehat{\alpha}(u)$ approaches $|\lambda_{\max}(J_F)|$ as $K_{\mathrm{pi}}\to\infty$; for a non-normal $J_F$, finite $K_{\mathrm{pi}}$ can reflect transient amplification and need not certify $\rho(J_F)$~\citep{trefethen2005spectra}.
The right-hand trace in the main-text figure plots $\widehat{\alpha}(u_k)$ at each iterate $u_k$ with $K_{\mathrm{pi}}{=}20$ (float32).
At FPRM's halt (step 36) the residual is $3.2\cdot10^{-2}$ and $\widehat{\alpha}\approx1.03$; after the halt the residual stays between $1.3\cdot10^{-2}$ and $3.1\cdot10^{-2}$.
GFPR's residual falls below $10^{-12}$ by step 27, with $\widehat{\alpha}\approx1.4\times10^{-3}$ at the final step of the trace.
On the final states of this one puzzle we cross-check with restarted Arnoldi on the flattened Jacobian (Krylov dimension 35, 3 restarts; exact JVPs for FPRM, finite-difference products for GFPR).
The Ritz values are not the full spectrum, but they approximate its outer part.
For FPRM near its halt (step 33 in this run) the largest Ritz values are the pair $-0.126\pm1.031i$ ($|\lambda|\approx1.039$), followed by real values near $-0.98$; no Ritz value has real part outside $(-1,1)$.
The pair becomes contracting for the damped map only when $s<2(1-\operatorname{Re}\lambda)/|1-\lambda|^2\approx0.966$; the step size at halt is at least $0.997^3\approx0.991$, giving $|1-s+s\lambda|\approx1.03$.
For GFPR the largest Ritz value is $1.34\cdot10^{-3}$.

The population audits use held-out validation puzzles.
Halting statistics come from 1000 puzzles run with the authors' loop up to 35000 steps (median halt at step 126, median step size at halt $0.965$).
The population audit applies the same probe on 128 puzzles after 8000 iterations in float32 with halting disabled, with $K_{\mathrm{pi}}{=}10$ at the final state; 75\% have $\widehat{\alpha}(u)>1$.
We have not run Arnoldi on these states, so this fraction is not a certified statement that $\rho(J_F)>1$ on 75\% of puzzles.

\end{document}